# Maze solving Algorithm for line following robot and derivation of linear path distance from nonlinear path

Shadman Sakib
Department of Naval Architecture & Marine Engineering
Bangladesh University of Engineering & Technology
Dhaka, Bangladesh
shadman.sakib_siam@yahoo.com

Anik Chowdhury, Shekh Tanvir Ahamed and Syed Imam Hasan
Department of Electrical & Electronic Engineering
Bangladesh University of Engineering & Technology
Dhaka, Bangladesh

*Abstract*— In this paper we have discussed a unique general algorithm for exploring and solving any kind of line maze with another simple one for simple mazes without loops or with loops having highest two branches none of which are inward. For the general algorithm, we need a method to map the whole maze, which is required if the maze is complex. The proposed maze mapping system is based on coordinate system and after mapping the whole maze as a graph in standard 'Adjacency-list representation' method, shortest path and shortest time path was extracted using Dijkstra's algorithm. In order to find the coordinates of the turning points and junctions, linear distances between the points are needed, for which wheel encoder was used. However, due to non-linear movement of robot, the directly measured distance from the encoder has some error and to remove this error an idea is built up which ended by deriving equations that give us almost exact linear distance between two points from the reading of wheel encoder of the robot moving in a non-linear path.

## I. Introduction

In mobile robotics, maze solving problem is one of the most common problems and to solve this problem an autonomous robot is used. Mazes can be of different kinds, like- without any loops, having loops, grid system or without grid system. Numerous methods and algorithms have been developed having their own merits and techniques such as flood fill algorithm [3], breadth first-search [4], Prim's algorithm [5] etc. But they all need to know the position of the goal. So, they are not appropriate for solving mazes where the goal is unknown until discovering.

If there is no loop and we do not need to find the shortest path, then we do not need to map the whole maze for reaching the destination. This holds still true for the maze having 'small loop'. Here, by 'small loop' we mean loop with no inward branches and there can be highest two paths for getting in or out of the loop. And the loop will not contain the end-point.

In this work we have introduced a new method of maze solving for any kind of mazes. In order to solve a complex line maze, the main problem arises is to map the whole maze. Without mapping the whole maze, robot cannot find the shortest path. So, our main target is to map the whole maze and then find the shortest path. Our first algorithm is to solve simple mazes fulfilling the criteria mentioned before. Actually the first algorithm is an upgraded version of the most common and ancient maze solving method- 'Follow-the-right' or 'Follow-the-left'. The second algorithm is designed to solve and find the shortest path for any kind of maze. We have mainly focused on the mapping method using coordinate system and direction, by which we can save the whole maze as a graph in standard 'Adjacency-list representation' method [5]. Using this method the robot will get all adjacent points of each point and their coordinates. Then it will generate the desired graph of the maze in G = (V, E) format. Wheel encoder is used to determine the distance and the coordinates of the junctions and the turning points. However using wheel encoder value directly may bring some error because of non-linear movement of robot. Hence, we also discussed about the process of linearization of the distance that is computed from the wheel encoder reading.

## II. 'No-loop' or 'Short loop' maze

In this algorithm, the robot is instructed to follow a preference of directions. There may be two sets of sequences of preference. They are- (1) 'Left-Front-Right-Down' and (2) 'Right-Front-Left-Down'. For example, we may mark directions as follow: (1) for right, (2) for up, (3) for left and (4) for down. Whenever a junction is encountered, the robot will always choose a path based on the preference sequence. Let, there are two paths coming out of one path, one goes to the left, another to the right. If the robot comes across it, it will first go to the left path (For sequence 1) to explore it. This junction and taken direction will be saved in the memory. If after following the path, it comes to a dead end it will turn back and come to the previous position. After reaching that position, previously saved value in the data structure will be extracted and the present preferred direction number will be added to that value. This new value will be saved again in the same place of the data structure. Same procedure will be followed continuously until it finds the 'end-point'. So, the robot will have a record of number of junctions it came across and a corresponding array to record the taken direction. Moving like this, one time the robot will reach the 'end-point' and the recursively updated direction array will generate an error free simple path from starting point to finishing point. Here, is an example showing how it works−

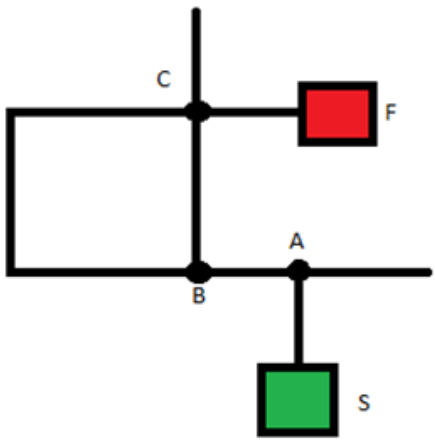

Fig. 1 An example of generalized algorithm for 'no-loop' or 'short loop' maze (showing only the points of interest)

For example we have a maze in the fig.1 (showing only the points of interest). The robot starts roaming from point 'S' and stops at point 'F'. We need an integer to keep record of present junction and an array to save the movement at each junction it came across. The preferred sequence of the robot is 'Right-Front-Left-Down'. Initially, present junction number and each elements of the array are 0. Now when the robot comes to junction 'A', at first it increases the present junction number by 1. Here it has a path at right (1) so it takes that path, add 1 to the current value of the 1$^{st}$ element of the array (As the present junction is 1), which is 0 and save it in that place. So, the value of the 1$^{st}$ element of the array will be 1. Following the line it will encounter a dead end and turn back. As it is turning back to the previous path, present junction will be decreased by 1 to 0. When it comes to the junction 'A' again, First, it will again increase present junction to 1. Now, it has two paths- one in the front (2) and another to the left (3). According to the preference, it will take the straight path and preferred direction number (2) will be added to the previously saved value in the 1$^{st}$ element of the array. Therefore its current value of that element will be 3. Now when it comes to junction 'B', present junction will be 2. Here it will find a path in the right (1) so it will take that path and add 1 to the value of the 2$^{nd}$ element of the array. So, value of the 2$^{nd}$ element of the array will be 1. When it comes to the junction 'C' same procedure will happen and eventually it will come to the 'end-point'. So after the exploration, the array will look like this – [3 1 1] (Indexing from 1). Hence, starting from the beginning when it comes to junction 'A' it will take the direction indicated in the first element of the array which is 3, thus left (3) and when it comes to the junction 'B' it will go right ('1' is for right) and at the junction 'C' it will again take the right ('1', for right) path and will come to the end-point, which is the shortest path of the explored paths by the robot.

### III. ALGORITHM FOR SOLVING ANY KIND OF MAZE

This solving technique is developed for the case where we do not have any grid system, there is no restriction about the loops and we have to find out the exact shortest path. It is said before that our main concern here is to map the whole maze and store it as a graph in standard 'Adjacency-list representation' method. In this method, we must know the adjacent points of each point. Then we can solve the graph for shortest path using Dijkstra's algorithm [2]. There are two types of points we need to know about- turning points and junction points.

In this algorithm, there will be four arrays. Among them one is two dimensional. This two dimensional array is used to store the coordinates (x, y) of the points. Rest arrays are vectors. They are –

*A. Point array* - saves each point by a distinct number or character in the serial they are explored.

*B. Type array* - identifies the type of the corresponding point of *Point array*.

*C. Explored array* - keeps record how many times a junction point is differently explored by the robot.

Elements of the *Type array* is one less than the number of ways a point directs towards. It actually indicates how many times we need to go through that point (each time choosing a different path) to explore all adjacent points of the respective point. Again, the number of paths we have already explored will be saved in the *Explored array*. As long as for each point the values of their corresponding element of *Explored array* and *Type array* are not same; we have to continue scouting to map the maze because there are still some ways or even points to be explored.

As mentioned earlier, there will be four arrays and two variables i.e. *Total_points* and *Direction*. At the beginning of the exploration the robot encounters the starting point. We name this point as '0'. So the first element of *Point array* will be 0 (*Point* [0] = 0) and the variable *Total_points* will be 1. The robot will also save the coordinate of that point in *Coordinate array* i.e. *Coordinate* [0] [0] = 0, *Coordinate* [0] [1] = 0. Here the first indexing element of the array represents the point number and the second one represents the x-coordinate or y-coordinate. The robot will save a point's x and y coordinates in its corresponding position in this 2D array. By doing this, it will get the point number and its corresponding coordinates. Now, it will take a direction and again the value of the variable *Direction* will change. In this algorithm, as we are to map the whole maze, absolute direction is used i.e. North-South-East-West. Say if it takes an eastern direction then the value of *Direction* will be (1), for north it will be (2), (3) will be for west and (4) for south. Values can be different, but increment must be in a cyclic order. Initial direction will be regarded as north. After taking a direction it will start to read the encoder value and follow the line until a junction or turning point is reached. After reaching a junction or turning point it will stop there and calculate the distance and identify coordinate. Now two situations may arise whether it can be a new point or an old one. In order to find whether it is a new point or an old one, we cross check the coordinates of that point with the elements of *Coordinate array* within a predefined tolerance value. If the point is a new one, then the robot will save it in the *Point array* with a new name, increases the *Total_Points* value, save type value in the *Type array*, save the ways it has been explored in the corresponding element for that point in the *Explored array* and also save coordinates of that point in *Coordinate array*. Then it will again begin to explore and take a direction from the sequence of preference. For exploration it will do the same thing again and again saving all kinds of values for each point i.e. *Point* number, *Type*, *Explored* number, *Coordinate* and *Direction*. If the point is not a new one then it will save the point again in the *Point array* with its previous name, find the first position of that point in the *Point array* and increase the value of its corresponding element of *Explored array* if last point does not exist before adjacently to this point in the *Point array*. Value of the element of *Explored array* for any corresponding point is equal to one less than the number of its explored total different adjacent points. If the *Explored* value is less than the *Type* value then it will take the unexplored path and explore it. If the *Explored* value of that point is equal to the *Type* value then it will search the *Point array* and find those points of which *Explored* value is less than *Type* value. If there is any point like this, then it will go to explore it by using the *Point array*, *Direction* and *Coordinate array*. If there is no point then the exploration is finished. By doing this recursively it will explore the whole maze and get all the junctions and turning points, their coordinates and their adjacent points, which are necessary to build the graph.

**Case study:** Here is a maze for demonstrating the exploring technique of the robot. (fig. 2) At first the robot starts from the starting zone, gives it a name in *Point array*('S'). (It is more convenient if we use integer to number the points serially like 1, 2, 3… instead of A, B, C… But here, for easy representation, we have used Characters) and identify the coordinate of this point as (0, 0) and save it in the *Coordinate array*. As it is a dead end, the *Type* value will be 0 and the *Explored* value will be 1. Now the robot is in the forward direction which will be considered as north. So the *Direction* value will be 2 as we assumed earlier and the *Total_Points* value will be 1. Now it will go to forward direction and start taking the encoder value. When it will reach a point, it will check the coordinates of that point. As the robot came from (0, 0) point to north side and say total distance is 10 units, that means it moved along the y-axis in positive direction. So the coordinate will be (0, 10). It does not match to the coordinates of any of the previously visited points. So it is a new point. As it is a four way junction, the value of *Type* will be 3 and the *Explored* value will be 1. Now suppose it prefers the eastern path. So, it will take a right turn and the value of *Direction* will be updated now. It will again begin to explore and after reaching the new point, it will identify its coordinate 'E' (14,10), of which *Type* value will be 2. Now it will again take the northern direction and find new point 'D' (14, 13). Similarly, it will explore the 'G' (14, 16) point. Here it will search for the points having smaller *Explored* number than their *Type* number and come back to 'D', as it is one of them and the closest one. Now, as this is an old point, it will find out the explored adjacent points of this point from the array which are 'G' and 'E'. Analyzing their coordinates it will know that they are in north and east side and the only available new path is the western one. So, it will go through that path. D's *Explored* number will be now 2; equal to the *Type* number. After a while, the robot will again reach 'A' point passing 'C'. Now, as it came here through a different path, *Explored* number will be increased. After analyzing the adjacent points' coordinates, western path will be chosen. *Explored* number will be again increased. So, it will be now 3; equal to the *Type* number. Eventually it will explore point 'B' which is a dead end. So, here, it will search for the points having smaller *Explored* number than their *Type* number. Only one such point exists − 'E'. So, it will go to the 'E' point and take the southern path, which leads to the dead-end end-point 'F'. Now, there is no points available having smaller *Explored* number than *Type* number. So, the exploration is finished.

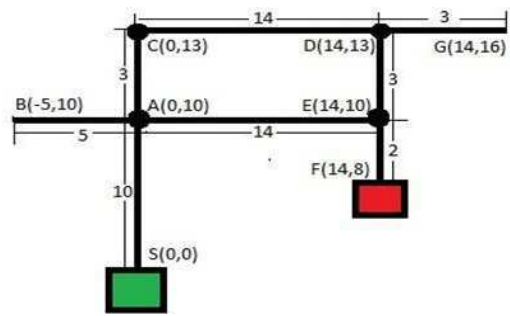

Fig. 2 An example of maze for proposed maze mapping algorithm for any kind of maze (with showing all the points and distance between points)

## IV. GENERATING GRAPH OF STANDARD FORMAT AND FINDING THE SHORTEST PATH

The last stage of our mapping algorithm is making an adjacency-list for each point from the obtained data. We can easily obtain this from the *Point array*. To do so, we will take one particular point from that array and mark each position of that point in the *Point array*. Then we will find out all points which are in the immediate next or previous position to that particular point. These are the adjacent points to that particular point. Doing this for each point will give us the complete 'Adjacency-list representation' of the graphed maze in G = (V, E) format. To determine the node to node distance, we will also need the coordinates of each point which we have already stored in the Coordinate array. After that, using A* or Dijkstra's or any other algorithm we can extract the shortest path between nodes. The following figure of the data structure will help to realize this process –

| Point | S | A | E | D | G | D | C | A | B | A | E | F |
|---|---|---|---|---|---|---|---|---|---|---|---|---|
| Type | 0 | 3 | 2 | 2 | 0 | 2 | 1 | 3 | 0 | 3 | 2 | 0 |
| Explored | 1 | 3 | 2 | 2 | 1 | 2 | 1 | 3 | 1 | 4 | 2 | 1 |
| CoOr_X | 0 | 0 | 14 | 14 | 14 | 14 | 0 | 0 | -5 | 0 | 14 | 14 |
| CoOr_y | 0 | 10 | 10 | 13 | 16 | 13 | 13 | 10 | 10 | 10 | 10 | 8 |

So, A ⟶ S, E, C, B

Fig. 3 Data structure of generating the whole maze as G = (V, E) format from the explored points of the maze with corresponding coordinates. (Coordinates are determined by the corresponding point to point distance obtained by using wheel encoder and the equations of distance which is derived in the later part of the paper with taking account some minor error. Generally the error is ±3%)

## V. OBTAINING PRECISE DISTANCE FROM WHEEL ENCODER

In grid less mazes, a robot must know its current position and calculate distance between nodes. Not only in maze solving, but also in many other fields, it is important to measure distance precisely using wheel encoder. However, while measuring the length of a straight line, wheel encoder cannot give us the exact value directly. This is due to the fact that a line following robot does not always go along the straight line. The path followed by the robot is rather zigzag. There are two reasons behind this behavior — (1) Imperfect alignment of robot with the line which causes linear deviation from the line. (2) Difference between the RPMs (Revolutions per minute) of the driving motors which causes circular deviation from the line.

Here, we inspected the reasons behind the curved path of the robot and derived equations to extract the linear distance between two points from the computed zigzag path distance.

### A. Initial Alignment of Robot

A robot does not remain in perfect alignment with the line when it starts moving. So, it goes in a straight line (AJ in the fig. 4) making an acute angle with the line to be followed (AB in the fig. 4), if the speeds of the driving motors are same. Line detecting sensors detect the deviation when a minimum distance ('h' in fig. 4) is created between the middle sensor (midpoint of the robot's front side) and the line.

Then, the robot makes a right turn making an angle of θ with the path. The robot will go through this new line until again it creates h distance from the line in opposite distance. As the distance is same, the rotation will be same as the previous one (θ). Because it is a conditional rotation, not time based. This will continue. So, the path followed by the robot is like the fig. 4.

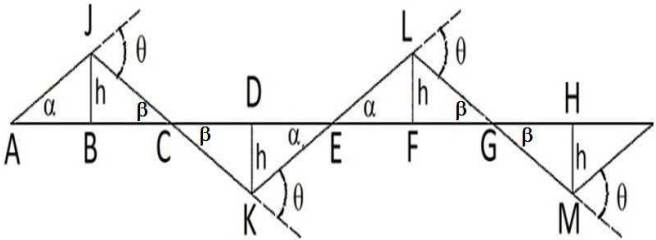

Fig.4 While following a linear path, robot deviates from the straight line if it is not fully aligned with the line. So, the robot goes to left side with an angle of $\alpha$ with the line to be followed until the sensors detects the deviation and again turns to the right side with an angle of $\theta$ with the line and the path the robot is moving looks like zigzag path.

In fig. 4, as rotation is same in points J and K, AJ and KL are parallel. Now, in ΔABJ, AB = AJ × *cosα*. Similarly, BC = CJ × *cosβ*, CD = CK × *cosβ*.

Adding them all together, we get-

AB + BC + CD + DE + EF + FG + GH = (AJ + KE + EL) × *cosα* + (JC + CK + LG + GM) × *cosβ*.

So, linear distance ≈ ($C_\beta$ × Total wheel distance after right turn) + ($C_\alpha$ × Total wheel distance after left turn)

Where, $C_\alpha$ = *cosα* and $C_\beta$ = *cosβ*.

However, measuring distance so frequently (after each rotational move) generates more error. To mitigate this problem, we have used one constant (*C*) instead of $C_\alpha$ and $C_\beta$. So, we can write-

AB + BC + CD + DE + EF + FG + GH = (AJ + JC + CK + KE + EL + LG + GM) × *C*

However, when a robot changes direction with rotational move, the wheels crosses some distance in a nonlinear, circular path. In fig. 5, left wheel of the robot goes through JD (straight line) until deviation reaches h, then it goes in an arc shaped path DE. Then again goes in a straight path EH. This goes beyond our assumption in fig. 4.

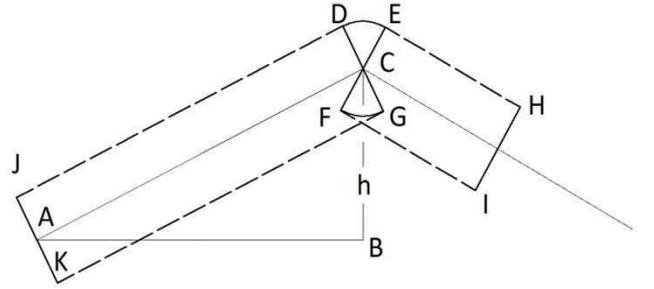

Fig. 5 When changing direction with rotational move, the wheel crosses some distance in nonlinear circular path.

So, we have to change our equation for extracting linear distance—

$D = [WL_{Total} - F_{LC} \times (N_R + N_L) - K \times N_L] \times C_L + F_{LL} \times N_R$ (For left wheel).  (1)

Or,

$D = [WL_{Total} - F_{RC} \times (N_R + N_L) - K \times N_R] \times C_R + F_{RL} \times N_L$ (For right wheel).  (2)

Where, $F_{RC}$ and $F_{RL}$ are respectively the circular and linear distance that the right wheel crosses while rotating.

$F_{LC}$ and $F_{LL}$ are respectively circular and linear distance that the right wheel crosses while rotating.

$N_R$ and $N_L$ are number of total right and left turns respectively.

$C_L$ and $C_R$ is the value of *C* for left and right wheel respectively.

$WL_{Total}$ and $WR_{Total}$ are total distance crossed by left wheel and right wheel respectively and K is inner rotational constant.

### B. Speed difference of the driving motor

Having motors of exact same RPM is unlikely. Even after using PWM (Pulse Width Modulation), there remains a little difference in RPM between two motors. This speed difference leads to deviation from straight line and causes curved path.

If we let the robot go freely where, $V_R$ (RPM of right motor)> $V_L$ (RPM of left motor), the robot will follow a circle.

If the radius of the circle travelled by the left wheel is $R'$ and $R''$ for the right wheel then as in same time the wheels travelled the perimeter of these circles, $V_R \times T = 2\pi R''$ and $V_L \times T = 2\pi R'$.

So, $\frac{V_R \times T}{V_L \times T} = \frac{2\pi R''}{2\pi R'}$ or, speed ratio $= \frac{R''}{R'} = 1 + \frac{B}{R'}$. (3)

Where, $B$ = Distance between two wheels.

So, the robot will always maintain same radius while moving in a circular or arc shaped path. This radius will increase if the difference between the RPM of the motors decreases. It is assumed that the RPM difference is very small. However, the path line of the robot among successive two rotational moves is not linear, rather circular. So, the path line of the robot will be similar to fig. 6. This means, the distance we get from the wheel encoder, is of an arc length rather than a straight line.

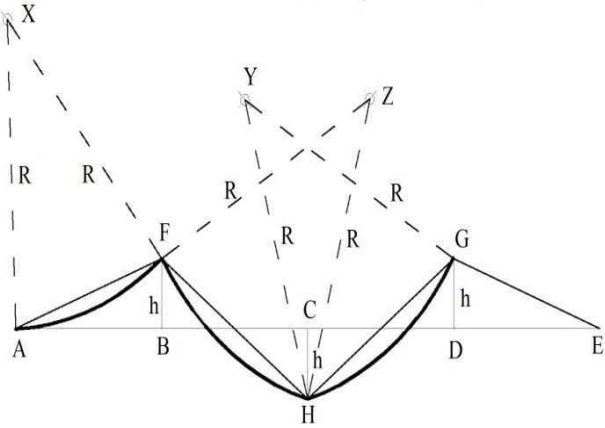

Fig. 6 As the RPMs of two motors are not identical the robot will follow the line not in a straight line rather in arc shaped line as the speed ratio (from equation 3) is not unity

If the equation of the circle be $x^2 + y^2 = R^2$, considering the first portion of fig. 6. Arc length,

$$ds = \int \sqrt{(dx)^2 + (dy)^2} \text{ or } s = \int_x^{x+h} \sqrt{1 + \left(\frac{dx}{dy}\right)^2} \, dy$$

Again, differentiating the equation of circle gives - $\frac{dx}{dy} = -\frac{y}{x}$. From this value, we get $s = \int_{x_1}^{x_2} \sqrt{(1 + \frac{x^2}{y^2})} \, dx = R \left[ sin^{-1}(\frac{x}{R}) \right]_{x_1}^{x_2} = R \left\{ sin^{-1}(\frac{\Delta x}{R}) \right\} [\Delta x = x_2 - x_1]$ or, $\Delta x = R \sin(\frac{s}{R})$.

This equation can extract length of the straight line from the length of the curved path the robot follows. However it can only be used when the path is a continuous arc. So, we have to read wheel encoder before each rotation and calculate the distance from last rotation point, which is not an efficient way and yields to very high error rate. So, we need to find out another way so that we can use the total travelled distance to find out the length of the straight line at once. From fig. 4 and fig. 7, we can see that the distance, a robot travels along the y-axis while moving in a continuous arc is of height either h or

2h unit. The arc length of an arc of height h can be calculated from this formula: $S_h = \int_y^{y+h} \sqrt{(1 + \frac{y^2}{x^2})} \, dy = R \sin^{-1}(\frac{h}{R})$.

For arcs having ordinate length of 2h, the equation of arc length becomes: $S_{2h} = R \sin^{-1}(\frac{2h}{R})$ (4)

Now, when robot goes in a zigzag way we considered the $[WL_{Total} - F_{LC} \times (N_R + N_L) - K \times N_L]$ term as a linear distance, which is not true if the RPMs of both motors are not same.

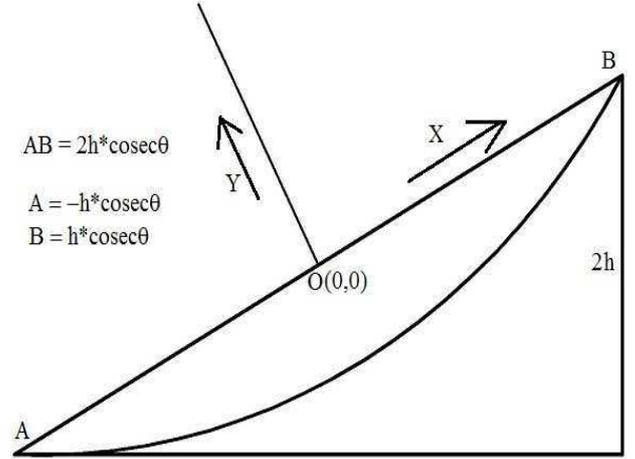

Fig. 7 The distance the robot travels along the y-axis while moving in a continuous arc is of height either h or 2h. Calculating the length of the arc path and corresponding x-axis direction mitigates the rate of error.

This is a sum of several arc lengths. Among them, $(N_R + N_L - 1)$ are of 2h height, 1 of h and another is not fixed. Now, we can modify that equation to get a more correct one. From fig.7 we can say, $s_{2h} = \int_{-h cosec\theta}^{h cosec\theta} \sqrt{(1 + \frac{x^2}{y^2})} \, dx = 2R \sin^{-1}(\frac{X_{2h}}{2R})$.

So, $X_{2h} = 2R \sin(\frac{S_{2h}}{2R})$ and $X_h = R \sin(\frac{S_h}{R})$ (5)

Arcs having 2h and h heights have a total length of $(N_R + N_L - 1) \times S_{2h} + S_h$.

So, arc length of last stage, $S_D = [WL_{Total} - F_{LC} \times (N_R + N_L) - K \times N_L] - (N_R + N_L - 1) \times S_{2h} - S_h$.

This is an arc which can be of various lengths. So,

$D_D = R \times \sin(\frac{S_D}{R})$. (6)

Now, we can find the equations for linear distance measurement-

$D = [(N_R + N_L - 1) \times X_{2h} + X_h + D_D] \times C_R + F_{RL} \times N_L$ (For right wheel). (7)

$D = [(N_R + N_L - 1) \times X_{2h} + X_h + D_D] \times C_L + F_{LL} \times N_R$ (For left wheel). (8)

These equations are the final equations which lead us to the exact length of a straight line from wheel encoder reading of a line follower robot. This equation also shows that if we know the minimum deviation for turn (h) and other constants ($C_R, F_{LL} ... etc$) with the total turn, we can predict the path distance to a fair approximation without even knowing the travelled distance. That means, for rough calculation of distance, wheel encoder is not needed at all.

## VI. RESULTS

Our proposed mapping technique along with Dijkstra's algorithm solves mazes of any shape. However mazes with large number of nodes need IC capable of high computational power. Again implementing the proposed equations we derived following data was found –

TABLE I. SAMPLE MEASURED PATH DISTANCE

| Type of Length | Test Number | | |
|---|---|---|---|
| | 1st test | 2nd test | 3rd test |
| Actual length (cm) | 10 | 14 | 8 |
| Length from wheel encoder (cm) | 10.23 | 14.14 | 7.90 |
| Length from formula (cm) | 10.02 | 14.04 | 7.99 |

If we include Kalman filter with this, error will be more less. So, we will get an very efficient system. But from fig. 8 we can see that the signal from wheel encoder is pretty much steady. So, filter is not necessary for the encoders. But it may be applied to the line following portion.

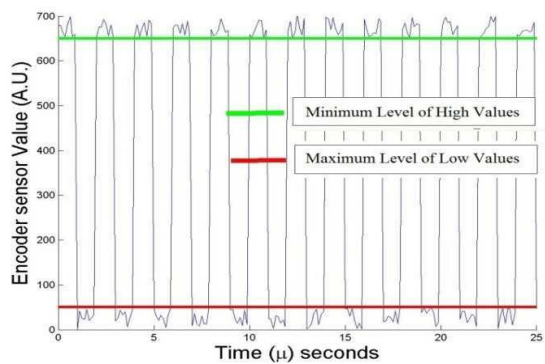

Fig. 8 The ADC (Analog to Digital Converter) value of wheel encoder suggests that further normalization is not needed for encoder value as the rate of error is negligible.

In many maze solving algorithms, it is important to know the position of the goal. So, they are not appropriate for solving line mazes where the goal is unknown until discovering. The proposed algorithm mitigates this limitation.

## VII. CONCLUSION

Different types of maze solving algorithm are often needed for different mazes. However there is no perfect general algorithm for that purpose. The algorithm discussed in this paper will give us a general method applicable for almost any kind of maze. The algorithm can also be used in sectors beyond line following. After some modifications, it can be used for autonomous road mapping, cave mapping, exploring and many other purposes. In this cases low cost GPS devices can be used instead of wheel encoder for tracking coordinates. Again we often need to determine the linear path distance using wheel encoder of a robot which brings error because of the nonlinear motion of the robot. Using the equations derived here, we can find it accurately.